 \documentclass[pmlr,twocolumn,10pt]{jmlr} % W&CP article

\usepackage{booktabs}
\usepackage{siunitx}
\usepackage{threeparttable}
\usepackage{multirow}
\usepackage{array}
\usepackage{enumitem}

\usepackage[switch]{lineno}

\theorembodyfont{\upshape}
\theoremheaderfont{\scshape}
\theorempostheader{:}
\theoremsep{\newline}

\usepackage{placeins}

 \title[FairTune]{FairTune: A Bias-Aware Fine-Tuning Framework Towards Fair Heart Rate Prediction from PPG}

\author{%
\Name{Lovely Yeswanth Panchumarthi} \Email{lpanch2@emory.edu}\\
\addr Computer Science, Emory University
\AND
\Name{Saurabh Kataria}\Email{skatar6@emory.edu}\\
\addr School of Nursing, Emory University
\AND
\Name{Yi Wu} \Email{yi.wu-1@ou.edu}\\
\addr Computer Science, University of Oklahoma
\AND
\Name{Xiao Hu} \Email{xhu40@emory.edu}\\
\addr School of Nursing, Emory University
\AND
\Name{Alex Fedorov} \Email{avfedor@emory.edu}\\
\addr School of Nursing, Emory University
\AND
\Name{Hyunjung Gloria Kwak} \Email{hkwak30@emory.edu}\\
\addr School of Nursing, Emory University
}

\begin{document}
\maketitle

\begin{abstract}
Foundation models pretrained on physiological data such as photoplethysmography (PPG) signals are increasingly used to improve heart rate (HR) prediction across diverse settings. Fine-tuning these models for local deployment is often seen as a practical and scalable strategy. However, its impact on demographic fairness—particularly under domain shifts—remains underexplored.
We fine-tune PPG-GPT — a transformer-based foundation model pretrained on intensive care unit (ICU) data — across three heterogeneous datasets (ICU, wearable, smartphone) and systematically evaluate the effects on HR prediction accuracy and gender fairness. While fine-tuning substantially reduces mean absolute error (up to 80\%), it can simultaneously widen fairness gaps, especially in larger models and under significant distributional characteristics shifts.

To address this, we introduce \textbf{FairTune}, a bias-aware fine-tuning framework in which we benchmark three mitigation strategies: class weighting based on inverse group frequency (IF), Group Distributionally Robust Optimization (GroupDRO), and adversarial debiasing (ADV). We find that IF and GroupDRO significantly reduce fairness gaps without compromising accuracy, with effectiveness varying by deployment domain. Representation analyses further reveal that mitigation techniques reshape internal embeddings to reduce demographic clustering. 
Our findings highlight that fairness does not emerge as a natural byproduct of fine-tuning and that explicit mitigation is essential for equitable deployment of physiological foundation models.
\end{abstract}
\vspace{-2.0ex}
\section{Introduction}
Heart rate (HR) monitoring is essential for detecting conditions such as sepsis and arrhythmias, where accurate and timely intervention is critical (\cite{Johnston2020HR, van2021vital,balaji2024improving, de2018heart}). However, significant disparities in monitoring accuracy across different patient demographics—particularly gender, age, and skin tone—pose substantial challenges to equitable clinical care: systematic underestimation in certain populations could delay critical interventions, while overestimation might trigger unnecessary alarms that burden healthcare systems and compromise patient care (\cite{Mastoris2023Remote,sjoding2020racial}). Thus, ensuring equitable HR monitoring across diverse patient demographics is not merely a technical consideration but a fundamental requirement for clinical safety, healthcare equity, and optimal patient outcomes in real-world healthcare environments.

Photoplethysmography (PPG)-based HR measurement has become standard across both clinical and consumer settings for continuous cardiovascular monitoring due to its non-invasive nature, real-time capability, and compatibility (\cite{lee2022real, xiao2024remote}). However, PPG signals exhibit complex variability due to physiological factors, environmental conditions, and device characteristics, posing challenges for robust prediction across diverse clinical environments (\cite{icenhower2025investigating, de2020sensors,patti2023exploring, nie2024review}). While machine learning approaches offer potential for learning complex patterns from PPG signals, yet they often inherit or exacerbate biases present in training data, leading to systematic prediction errors across different patient populations (\cite{dogra2024shortcut, boland2024all, dehdashtian2024fairness}). Although fine-tuning a pretrained model on a new dataset may improve overall performance, it does not reliably mitigate biases embedded during pretraining or amplified downstream by skewed datasets (\cite{wang2023overwriting, gan2024erasing, zhang2024bias}).

Previous research in clinical AI has addressed domain adaptation, subgroup-aware modeling, and bias mitigation techniques such as reweighting (\cite{hainmueller2012entropy, krasanakis2018adaptive}), Group Distributionally Robust Optimization (GroupDRO) (\cite{sagawa2019distributionally, zhang2023stochastic}), and adversarial debiasing (ADV) (\cite{zhang2018mitigating}) in supervised learning. However, their impact on fine-tuning pretrained foundation models remains unclear. Foundation models introduce unique challenges during adaptation, as their large-scale pretraining can embed complex spurious correlations that may amplify biases depending on the downstream setting (\cite{bommasani2021opportunities, he2024foundation}). This issue is particularly critical in physiological domains such as PPG, where signal characteristics and demographic factors interact in complex ways. To our knowledge, no prior work has systematically examined whether fairness-aware fine-tuning can mitigate pre-existing or newly introduced biases in foundation models for physiological prediction tasks (\cite{chen2023multimodal, ding2024siamquality}). 

Fine-tuning enables practical local adaptation of foundation models to new populations and devices, particularly in settings where large-scale pretraining is not feasible (\cite{chen2024personalized, abnar2021exploring}). This approach allows models trained in data-rich environments to be effectively deployed in resource-limited contexts facing significant distribution shifts. In such settings, sharing pretrained models — rather than sensitive patient-level data — combined with lightweight local fine-tuning offers a feasible and privacy-preserving pathway for broader deployment of physiological AI technologies (\cite{baker2023artificial}). Nevertheless, models pretrained on rich datasets from one population — such as U.S. intensive care unit (ICU) data — may perform poorly when deployed across different demographic groups, healthcare systems, or wearable devices, due to shifts in data distribution. Thus, a bias-aware fine-tuning process that considers both performance and fairness across distribution shifts is essential for practical clinical deployment.

Many real-world clinical and public health settings face significant resource constraints that limit the availability of large, diverse, and well-labeled PPG datasets. Privacy regulations and logistical challenges often make data sharing infeasible, leading to increased reliance on model sharing as an alternative (\cite{bonomi2020privacy, ranganathan2002improving}). In these environments, the ability to fine-tune a foundation model with limited local data — while maintaining both generalization and fairness — rather than performing computationally expensive training from scratch, becomes critical for practical deployment (\cite{zhuang2023foundation}).

\begin{figure}[t] %ht
  \centering
  \includegraphics[width=1.0\linewidth]{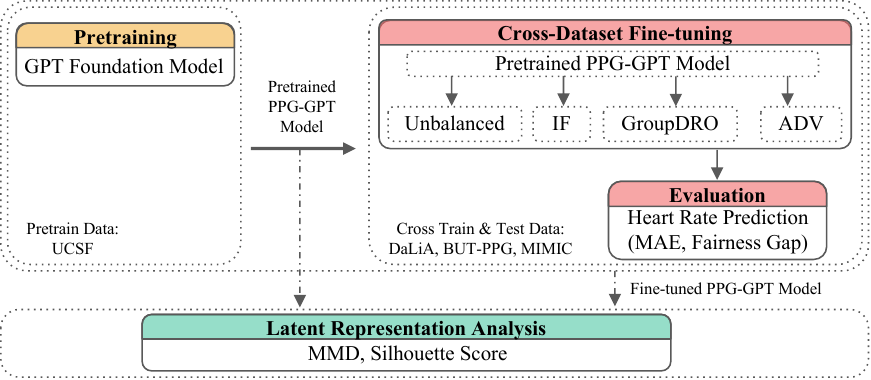}
  \caption{Overview of \textbf{FairTune}: A Bias-Aware Fine-Tuning Framework for PPG-GPT in HR Prediction.}
  \label{fig:figure1_overview}  
\end{figure}

To address these challenges in a practical and scalable manner, we propose \textbf{FairTune}, a bias-aware fine-tuning framework for physiological foundation models. Specifically, we:
\begin{itemize}
  \item Fine-tune a large pretrained PPG foundation model (PPG-GPT) for HR prediction across heterogeneous datasets under cross-dataset transfer scenarios.
  \item Evaluate bias-aware fine-tuning strategies, including class weighting based on inverse group frequency (IF), GroupDRO, and ADV.
  \item Analyze performance improvements and fairness trade-offs using quantitative metrics (Mean Absolute Error [MAE], fairness gaps) and latent representation metrics (Maximum Mean Discrepancy [MMD], Silhouette Score).
\end{itemize}
%Our findings reveal that bias-aware fine-tuning substantially improves both HR prediction accuracy and demographic fairness, highlighting its potential to bridge critical gaps in physiological AI and to facilitate the deployment of foundation models in diverse, resource-limited healthcare environments. % (see Figure~\ref{fig:figure1_overview} for a schematic overview).
Our findings reveal that bias-aware fine-tuning substantially improves both HR prediction accuracy and demographic fairness, highlighting its potential to bridge critical gaps in physiological AI and to facilitate deployment in diverse, resource-limited settings. Figure~\ref{fig:figure1_overview} provides an overview of our proposed framework, \textbf{FairTune}.
\vspace{-1.0ex}

\section{Methods}
%An overview of our proposed framework, \textbf{FairTune}, for bias-aware fine-tuning of PPG-GPT in HR prediction is shown in Fig. ~\ref{fig:figure1_overview}.%This includes foundation model pretraining, cross-domain fine-tuning with fairness-aware objectives, and multi-dimensional evaluation of prediction accuracy and demographic fairness.
\subsection{Pretrained Model: PPG-GPT}
We build upon \textbf{PPG-GPT}, a large-scale transformer-based foundation model pretrained on over 200M PPG waveform segments collected primarily from critical care settings to learn general-purpose physiological representations in a self-supervised fashion (\cite{chen2024adapting, chen2025gpt}). The model adopts a decoder-only transformer architecture that processes raw PPG signals through key components including a patch embedding layer that segments continuous waveforms into 40-sample patches, multi-head self-attention mechanisms with rotary positional embeddings, and a reconstruction head trained using Logit-Laplace loss to capture complex temporal dependencies and physiological patterns. 
%The model is a decoder-only transformer that ingests raw PPG via 40-sample patch embeddings and multi-head self-attention with rotary positional embeddings; a reconstruction head is trained with Logit-Laplace loss.
We experimented with four model sizes (19M-1B parameters) released in the original PPG-GPT to examine model complexity and its impact on accuracy–fairness trade-offs. We report results with the 19M configuration as the representative model for clinical adaptation. In this study, we initialize our HR prediction model with the pretrained PPG-GPT and apply bias-aware fine-tuning strategies to ensure equitable performance across diverse patient populations, without modifying the underlying architecture or pretraining weights.%We experimented with four model sizes (19M-1B parameters) released in the original PPG-GPT to examine model complexity and its impact on accuracy–fairness trade-offs. In the main text, we report results with the 19M configuration as the representative model for clinical adaptation, while complete scaling analyses across all parameter sizes and dataset combinations are provided in Appendix \ref{apd:second}.
%In this study, we initialize our HR prediction model with the pretrained PPG-GPT and apply bias-aware fine-tuning strategies to ensure equitable performance across diverse patient populations, without modifying the underlying architecture or pretraining weights.
\begin{table*}[ht]
\floatconts
  {tab:gender_distribution}%
  {\caption{Demographic Distribution by Gender across Datasets.}}%
  {%
    \small
    \begin{threeparttable}
    \resizebox{\linewidth}{!}{%
    \begin{tabular}{l|cc|cc|cc}
        \toprule
        \multirow{3}{*}{\textbf{Measure}} & \multicolumn{4}{c|}{\textbf{Consumer}} & \multicolumn{2}{c}{\textbf{Clinical}} \\ 
        \cmidrule(lr){2-5} \cmidrule(l){6-7}
        & \multicolumn{2}{c|}{\textbf{DaLiA}} & \multicolumn{2}{c|}{\textbf{BUT-PPG}} & \multicolumn{2}{c}{\textbf{MIMIC}} \\ 
        \cmidrule(lr){2-3} \cmidrule(lr){4-5} \cmidrule(l){6-7}
        & \textbf{Female} & \textbf{Male} & \textbf{Female} & \textbf{Male} & \textbf{Female} & \textbf{Male} \\ 
        \midrule
        Sample Size, n (\%) & \textbf{35,099 (54.3\%)} & 29,598 (45.7\%) & \textbf{1,986 (51.1\%)} & 1,902 (48.9\%) & \textbf{6,804 (62.3\%)} & 4,123 (37.7\%) \\ 
        \midrule
        Age (yrs), med [IQR] & 24.0 [21.0-26.0] & 28.0 [25.0-43.0] & 23.0 [22.0-49.0] & 25.0 [22.0-43.0] & 50.0 [25.0-72.0] & 46.0 [23.0-69.0] \\ 
        \midrule
        HR (bpm), med [IQR] & 85.2 [73.3-96.8] & 85.5 [73.4-110.2] & 75.0 [70.0-83.0] & 78.0 [67.0-87.0] & 88.8 [74.2-105.6] & 90.3 [75.1-102.5] \\ 
        \bottomrule
    \end{tabular}%
    }%
    \end{threeparttable}
  }
\end{table*}
\vspace{-1.0ex}
\subsection{Datasets}
We used three real-world PPG datasets that span two major acquisition settings: consumer and clinical. 
The consumer datasets include \textbf{PPG-DaLiA} and \textbf{BUT-PPG}, which differ in device type and recording context (\cite{reiss2019deep, nemcova2021brno}). 
DaLiA contains PPG signals recorded at 64Hz from wrist-worn sensors during daily activities, including motion-rich scenarios (e.g., walking, working). In contrast, BUT-PPG provides 10-second smartphone-based recordings captured with a Xiaomi Mi9 using a fingertip-on-camera setup at 30Hz. The clinical dataset, \textbf{MIMIC} comprises high-resolution (125Hz) clinical PPG waveforms collected from ICU monitors, representing a high-stakes clinical environment (\cite{charlton2022detecting}). 
Table~\ref{tab:gender_distribution} summarizes the distribution of sample sizes, gender, age, and HR across datasets. To ensure comparability with varying sampling rates, we resampled all PPG signals to 40Hz, segmented them into 1-second non-overlapping windows, and applied basic signal-level imputation for missing values. Detailed pre-processing steps and dataset documentation will be provided at GitHub. %\href{https://github.com/ucabhkw/FairTune}{GitHub}.
\vspace{-1.0ex}
\subsection{Fine-tuning Procedure}
We fine-tuned PPG-GPT for HR prediction as a regression task using a composite loss function consisting of: (1) L1 Loss (MAE) as the primary regression objective for HR prediction, and (2) Logit Laplace Loss as an auxiliary regularization term to improve the model's distributional understanding of PPG signals. Full layer fine-tuning was performed separately for each dataset and in cross-dataset settings.
We used Adam optimization with carefully tuned hyperparameters ($\beta_1$=0.9, $\beta_2$=0.999, weight decay=$1e^{-5}$) and implemented a sophisticated learning rate with 10\% of warm-up training steps (from $1e^{-5}$ to $10e^{-5}$), followed by cosine annealing decay to $1e^{-6}$. Input waveforms were standardized, and the fine-tuning was performed for up to 50 epochs with mini-batches. 

\vspace{-1.0ex}
\subsection{Bias Mitigation Strategies}
%To promote fairness across gender groups, We explored three bias mitigation strategies during fine-tuning
We explored three bias mitigation strategies during fine-tuning:
\begin{itemize}
  \item Class Weighting (Inverse-Frequency Weighting, IF): Samples were weighted inversely to group frequency, assigning larger weights to underrepresented groups in the training set.
  %We implemented sample weighting inversely proportional to gender frequency in the training data. This approach assigns higher importance to underrepresented groups, effectively balancing the contribution of each demographic during optimization. %The implementation creates weight coefficients and incorporates them through a weighted sampler to ensure proportional representation during batch generation. Throughout the manuscript, we refer to this strategy as ``Inverse-Frequency Weighting (IF)'' in Figures and Tables.
  \item Group Distributionally Robust Optimization (GroupDRO): Group-specific losses were tracked separately, and weights were updated dynamically so that optimization emphasized the group with the highest loss.%Our implementation tracked gender-specific losses separately and dynamically updates group weights during training using an exponential moving average approach. By weighting the objective function toward the group with the highest error, GroupDRO ensures that optimization focuses on reducing disparities rather than just minimizing average error.
  \item Adversarial Debiasing (ADV):   A gradient-reversal adversary was trained to predict group labels from intermediate representations while the main model minimized prediction error and maximized the adversary’s loss.
  %We employed a gradient reversal approach where a secondary neural network (adversary) attempts to predict gender from model embeddings while the primary model is trained to simultaneously minimize heart rate prediction error and maximize the adversary's uncertainty. This creates competing objectives: the adversary becomes better at detecting gender information in representations, while the primary model learns to remove gender-revealing features. %We implemented this using an entropy-based adversarial loss that encourages uniform gender prediction probabilities, effectively removing demographic information from internal representations.
\end{itemize}
Full mathematical formulations are provided in Appendix~\ref{apd:first}.
\vspace{-2.0ex}
\subsection{Cross-Dataset Evaluation}
To evaluate the generalizability and fairness of bias-aware fine-tuning, we conducted systematic cross-dataset experiments using all pairwise combinations of three PPG datasets. Starting from a PPG-GPT model pretrained on UCSF ICU data, we fine-tuned the model on one dataset (source) and evaluated it on a different dataset (target), without accessing target data during training.

We defined three evaluation categories: (1) \textbf{Consumer-to-Clinical:} DaLiA, BUT-PPG $\rightarrow$ MIMIC, (2) \textbf{Clinical-to-Consumer:} MIMIC $\rightarrow$ DaLiA, BUT-PPG and (3) \textbf{Cross-Consumer:} DaLiA $\leftrightarrow$ BUT-PPG.

For each source–target pair, we applied three bias mitigation strategies during fine-tuning and evaluated both predictive accuracy and demographic fairness on the target domain. This setup simulates realistic distribution shifts encountered when deploying foundation models across heterogeneous devices, populations, and care settings.
\vspace{-1.0ex}
\begin{table}[ht]
\floatconts
  {tab:pretrain_vs_finetune}%
  {\caption{Comparison of Pretrained and Fine-Tuned Model Performance across Datasets.}}%
  {%
    \small
    \begin{threeparttable}
    \resizebox{\linewidth}{!}{
    \begin{tabular}{l|c c|c c|c c}
    \toprule
    \multirow{2}{*}{\textbf{Dataset}} &
    \multicolumn{2}{c|}{\textbf{MAE}} &
    \multicolumn{2}{c|}{\textbf{Silh (True)}} &
    \multicolumn{2}{c}{\textbf{Silh (Pred)}} \\
    \cmidrule(lr){2-3} \cmidrule(lr){4-5} \cmidrule(l){6-7}
    {}& \textbf{Pre} & \textbf{Fine} & \textbf{Pre} & \textbf{Fine} & \textbf{Pre} & \textbf{Fine} \\
    \midrule
    DaLiA    & 89.54 & 17.62 & -0.0044 &  0.0167 & --  & 0.0203 \\
    BUT‑PPG  & 78.22 & 19.23 & -0.0458 & -0.0639 & --  & 0.5871 \\
    MIMIC    & 93.81 & 24.73 & -0.0105 & -0.0182 & --  & 0.6356 \\
    \bottomrule 
    \end{tabular}
    }
    \begin{minipage}{\linewidth}
    \scriptsize
    Silh; Silhouette, Pre; Pretrained, Fine; Fine-tuned
    \end{minipage}
    \end{threeparttable}
  }
\end{table}\vspace{-2.0ex}
\subsection{Evaluation Metrics} 
To comprehensively assess both predictive performance and demographic fairness of fine-tuned PPG-GPT models, we use:
\paragraph{Mean Absolute Error (MAE): } Measures average absolute difference between predicted and ground-truth HRs across the evaluation dataset. It provides a clinically interpretable metric where lower values indicate more accurate HR estimation. 
\paragraph{Fairness Gap: } Absolute difference in MAE between gender subgroups, indicating prediction disparity. Smaller values reflect better fairness.
\paragraph{Latent Representation Metrics: }
To evaluate whether the model encodes demographic bias over physiological features, we evaluate its internal representations using the following techniques:
%To evaluate whether the model's internal decision-making process relies on demographic shortcuts rather than physiologically relevant features, we analyze the learned representations using complementary techniques:

\begin{itemize}[leftmargin=1em, topsep=0pt, itemsep=1pt]
    \item Silhouette Score: Measures how well the model organizes patients in its internal feature space. We examine two aspects to understand different types of learned clustering:
    \begin{itemize}[noitemsep, topsep=0pt, leftmargin=*]
        \item \textit{Silhouette (True):} Evaluates whether patients with similar actual HRs are grouped in the model's internal representation, indicating clinically meaningful feature organization.
        \item \textit{Silhouette (Pred):} Assesses whether patients with similar predicted HRs cluster together, revealing the model's internal decision boundaries.    
    \end{itemize}
    HRs were grouped as bradycardia ($<$75), normal (75$-$95), and tachycardia ($>$95 bpm). Higher scores denote clearer physiological separation.%For this analysis, HRs are categorized into clinically relevant ranges: bradycardia ($<$75 bpm), normal (75--95 bpm), and tachycardia ($>$95 bpm). Higher scores indicate clearer physiological separation.
    \item Maximum Mean Discrepancy (MMD): 
    Measures distributional differences in final hidden layer embeddings between gender groups; lower values suggest reduced gender-specific representational disparity, indicating more uniform processing across genders (More details in Appendix ~\ref{apd:third}).
\end{itemize}
\begin{figure*}[!t]
  \centering
  \includegraphics[width=1\linewidth]{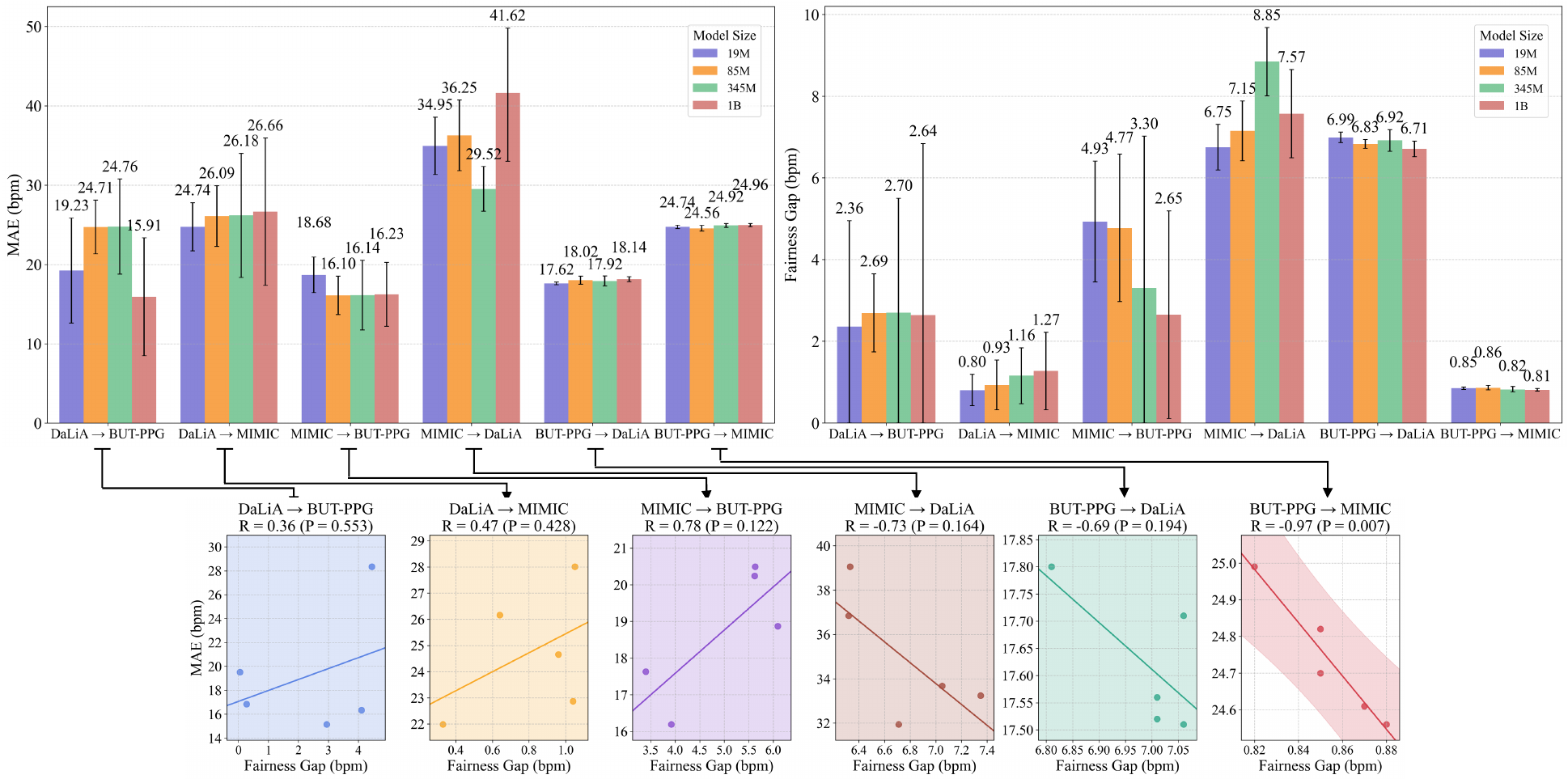} %model_complexity_fairness_2.pdf}
  \caption{Effect of PPG-GPT Model Scaling on HR Prediction Performance and Fairness. Top: MAE and fairness gap trends across model sizes in six transfer scenarios. Bottom: Accuracy–fairness trade-offs for the 19M model across five random seeds.}%Each point represents one training run; the trend line summarizes the relationship between MAE and fairness gap per transfer direction.
  \label{fig:fairness_gap_unbalanced}
\end{figure*}
\vspace{-1ex}\begin{table*}[!t]
\floatconts
  {tab:Bias_Mitigation_Performance}%
  {\caption{Cross-Dataset Heart Rate Prediction Performance with Bias Mitigation Strategies for 19M PPG-GPT Model.}}%
  {%
    \small
    \begin{threeparttable}
    \resizebox{\linewidth}{!}{
    \begin{tabular}{ll|l|c|cc|c}
    \toprule
    \multirow{2}{*}{\textbf{Trainset}} & \multirow{2}{*}{\textbf{Testset}} & \multirow{2}{*}{\textbf{Method}} 
    & \multicolumn{3}{c|}{\textbf{Predictive Performance (MAE ↓)}} 
    & \textbf{Fairness Gap} \\
     & & & \textbf{Total} & \textbf{Male} & \textbf{Female} & \textbf{$|$Male$-$Female$|$} \\
    \midrule
    \multirow{4}{*}{DaLiA} & \multirow{4}{*}{BUT-PPG} & Unbalanced 
    & 19.23 [12.61--25.85] & 20.64 [13.84--27.44] & 17.90 [11.52--24.28] & 2.36 [--0.23--4.95] \\
    \cmidrule{3-7}
    & & IF 
    & \textbf{13.44 [11.52--15.36]} & 14.10 [12.86--15.34] & 12.80 [10.21--15.39] & 1.30 [--0.14--2.74] \\
    & & GroupDRO 
    & 15.47 [13.79--17.16] & 15.81 [13.83--17.79] & 15.15 [13.79--16.51] & \textbf{0.68 [--0.10--1.46]} \\
    & & ADV 
    & 14.49 [11.21--17.79] & 14.85 [11.28--18.42] & 14.17 [10.11--18.22] & 0.70 [--0.29--1.65] \\
    \midrule
    \multirow{4}{*}{DaLiA} & \multirow{4}{*}{MIMIC} & Unbalanced 
    & 24.74 [21.71--27.77] & 24.32 [21.57--27.07] & 24.99 [21.83--28.15] & 0.80 [0.42--1.19] \\
    \cmidrule{3-7}
    & & IF 
    & 22.72 [21.54--23.91] & 22.07 [20.78--23.36] & 23.13 [22.00--24.26] & 1.06 [0.91--1.21] \\
    & & GroupDRO 
    & 22.46 [21.05--23.86] & 21.71 [20.20--23.22] & 22.91 [21.62--24.20] & 1.20 [0.90--1.50] \\
    & & ADV 
    & 22.70 [20.05--25.35] & 21.98 [19.06--24.89] & 23.13 [20.63--25.64] & 1.16 [0.71--1.60] \\
    \midrule
    \multirow{4}{*}{BUT-PPG} & \multirow{4}{*}{DaLiA} & Unbalanced 
    & 17.62 [17.46--17.78] & 21.41 [21.30--21.52] & 14.42 [14.23--14.61] & 6.99 [6.86--7.12] \\
    \cmidrule{3-7}
    & & IF 
    & \textbf{17.65 [17.38--17.92]} & 21.37 [21.10--21.64] & 14.51 [14.25--14.77] & \textbf{6.86 [6.81--6.91]} \\
    & & GroupDRO 
    & 18.18 [17.83--18.53] & 22.04 [21.66--22.42] & 14.92 [14.66--15.18] & 7.12 [7.01--7.24] \\
    & & ADV 
    & 19.47 [18.62--20.31] & 23.35 [22.44--24.25] & 16.19 [15.39--16.99] & 7.15 [6.87--7.44] \\
    \midrule
    \multirow{4}{*}{BUT-PPG} & \multirow{4}{*}{MIMIC} & Unbalanced 
    & 24.74 [24.52--24.95] & 24.21 [23.85--24.57] & 25.06 [24.73--25.39] & 0.85 [0.82--0.88] \\
    \cmidrule{3-7}
    & & IF 
    & 24.62 [24.37--24.88] & 24.09 [23.81--24.37] & 24.95 [24.71--25.19] & 0.86 [0.82--0.89] \\
    & & GroupDRO 
    & \textbf{20.14 [20.01--20.28]} & 18.96 [18.74--19.18] & 20.86 [20.78--20.94] & 1.90 [1.74--2.06] \\
    & & ADV 
    & 24.92 [24.10--25.74] & 24.36 [23.51--25.22] & 25.26 [24.50--26.01] & \textbf{0.89 [0.79--1.00]} \\
    \midrule
    \multirow{4}{*}{MIMIC} & \multirow{4}{*}{BUT-PPG} & Unbalanced 
    & 18.68 [16.44--20.93] & 16.17 [14.50--17.84] & 21.10 [18.32--23.88] & 4.93 [3.45--6.41] \\
    \cmidrule{3-7}
    & & IF 
    & 29.64 [26.39--32.89] & 27.72 [24.10--31.34] & 31.48 [28.42--34.54] & 3.76 [2.73--4.80] \\
    & & GroupDRO 
    & 37.02 [29.09--44.95] & 36.14 [26.82--45.46] & 37.86 [30.81--44.91] & \textbf{2.03 [0.54--3.52]} \\
    & & ADV 
    & \textbf{27.21 [20.34--34.08]} & 24.73 [17.18--32.29] & 29.58 [23.45--35.72] & 4.85 [3.38--6.32] \\
    \midrule
    \multirow{4}{*}{MIMIC} & \multirow{4}{*}{DaLiA} & Unbalanced 
    & 34.95 [31.33--38.57] & 38.61 [35.20--42.02] & 31.86 [28.00--35.72] & 6.75 [6.19--7.31] \\
    \cmidrule{3-7}
    & & IF 
    & \textbf{25.24 [22.26--28.23]} & 29.36 [26.54--32.18] & 21.77 [18.67--24.87] & 7.59 [7.24--7.94] \\
    & & GroupDRO 
    & 42.15 [39.30--44.99] & 46.12 [43.65--48.59] & 38.79 [35.62--41.96] & 7.33 [6.64--8.02] \\
    & & ADV 
    & 36.54 [34.12--38.95] & 40.88 [38.50--43.27] & 32.87 [30.49--35.25] & \textbf{8.01 [7.38--8.64]} \\
    \bottomrule
    \end{tabular}
    }
    \begin{minipage}{\linewidth}
    \scriptsize
    Values are reported as median [95\% confidence interval] unless otherwise indicated. IF = Instance-weighted Fairness (Balanced training approach). Best performance values are shown in \textbf{bold} for each dataset pair.
    \end{minipage}
    \end{threeparttable}
  }
\end{table*}\vspace{-1ex}
\section{Results} \label{results}
\subsection{Impact of Fine-Tuning on Cross-Dataset HR Prediction}
\label{results_sub1}
Table~\ref{tab:pretrain_vs_finetune} compares the pretrained and fine-tuned (unbalanced) models trained on each target dataset. Fine-tuning PPG-GPT on context-specific datasets substantially improved HR prediction accuracy compared to the pretrained model without adaptation, reducing MAE substantially across all datasets: from 89.54 to 17.62 bpm for DaLiA (80\% reduction), from 78.22 to 19.23 bpm for BUT-PPG (75\%), and from 93.81 to 24.73 bpm for MIMIC (74\%). For reference, intra-dataset baselines achieved lower errors (see Appendix \ref{apd:fourth} Table ~\ref{tab:Intra_Bias_Mitigation_Performance}), confirming that the higher MAEs in cross-dataset settings primarily reflect domain shift rather than inherent model limitations .%These results demonstrate that the elevated MAE scores observed in cross-dataset scenarios (17-35 bpm) primarily reflect domain shift challenges rather than inherent model limitations.

Silhouette scores in Table~\ref{tab:pretrain_vs_finetune} quantify the degree of cluster separation in the model's latent space with respect to HR categories ($<$75 bpm, 75--95 bpm, $>$95 bpm). 
These scores assess how well the model's internal representations organize patients according to their physiological HR states, providing insight into whether the learned features capture clinically meaningful distinctions between different cardiac conditions.
For the pretrained model, the ``Silhouette (True)'' scores are slightly negative across all datasets (-0.0044 for DaLiA, -0.0458 for BUT-PPG, and -0.0105 for MIMIC), indicating minimal separation of embeddings according to true HR categories. After fine-tuning, we observe modest changes in the ``Silhouette (True)'' scores, with DaLiA showing a slight improvement to 0.0167, suggesting marginally better alignment between the embeddings and true HR categories.

\subsection{Model Complexity and Fairness Trade-offs} 
We evaluated the effect of scaling the PPG-GPT model size from 19M to 1B parameters on HR prediction accuracy (MAE, top right) and gender fairness gap (MAE difference between gender subgroups, top left) across different dataset transfers under an unbalanced setting without demographic reweighting (Fig. \ref{fig:fairness_gap_unbalanced}). The effects of scaling were highly heterogeneous, varying by dataset combination. The bottom panel shows the stability of accuracy-fairness trade-offs across multiple runs with random seeds for the 19M model.
\paragraph{Consumer-to-Clinical Adaptation: } 
In DaLiA $\rightarrow$ MIMIC, both overall MAE and fairness gap deteriorated with model scaling: MAE rose from 24.74 bpm (19M) to 26.66 bpm (1B), and the fairness gap increased from 0.80 to 1.27 bpm. In BUT-PPG $\rightarrow$ MIMIC, MAE remained stable across all model sizes (24.74–24.96 bpm) with fairness gap showing only marginal changes from 0.85 to 0.81 bpm.
\paragraph{Clinical-to-Consumer Adaptation: } 
Smartphone deployment (MIMIC $\rightarrow$ BUT-PPG) benefited from scaling with improved accuracy from 18.68 to 16.23 bpm and reduced fairness gaps from 4.93 to 2.65 bpm. However, wearable deployment (MIMIC $\rightarrow$ DaLiA) exhibited more varied patterns. The lowest MAE was observed with the 345M model (29.52 bpm), while the highest fairness gap also occurred at this size (8.85 bpm). At 1B parameters, MAE increased to 41.62 bpm and the fairness gap decreased slightly to 7.57 bpm.
%\vspace{-2.0ex}
\paragraph{Cross-Consumer Adaptation: } 
In DaLiA $\rightarrow$ BUT-PPG, MAE increased from 19.23 bpm (19M) to 24.76 bpm (345M), but dropped to 15.91 bpm with the largest model (1B). However, the fairness gap consistently increased from 2.36 bpm to 2.64 bpm. In BUT-PPG $\rightarrow$ DaLiA, Model scaling had little effect on overall accuracy, with MAE ranging from 17.62 to 18.14 bpm, and led to modest fairness improvements. The fairness gap decreased from 6.99 to 6.71 bpm. %Overall, larger models did not guarantee better performance, and in several cases they amplified demographic disparities. 

\subsection{Impact of Bias Mitigation Strategies}
We evaluated the impact of three bias mitigation strategies—IF, GroupDRO, and ADV—consistently reduced fairness gaps compared to unbalanced fine-tuning across all dataset transfers. Table~\ref{tab:Bias_Mitigation_Performance} summarizes the results. Across all transfer settings, IF achieved the best overall balance between accuracy and fairness. 
However, each method's effectiveness varied by source–target pair: IF was the most consistently effective, while GroupDRO and ADV had specific strengths in certain domains. These results suggest that bias mitigation strategies should be chosen on dataset-specific characteristics and compatibility between source and target data contexts. 

%\paragraph{Intra-Dataset Baseline Performance: }
%Within-dataset evaluation confirmed the effectiveness of bias mitigation strategies without cross-domain challenges. In DaLiA $\rightarrow$ DaLiA, IF achieved the best fairness improvement, reducing the gap from 0.49 to 0.30 bpm while maintaining accuracy (MAE: 4.28 bpm). For BUT-PPG $\rightarrow$ BUT-PPG, GroupDRO dramatically reduced the fairness gap from 3.15 to 0.18 bpm with minimal accuracy cost (MAE: 10.87 vs. 9.94 bpm). In MIMIC $\rightarrow$ MIMIC, GroupDRO again showed superior fairness performance, reducing the gap from 2.62 to 1.32 bpm.

\paragraph{Consumer-to-Clinical Adaptation: }
In DaLiA $\rightarrow$ MIMIC, all three strategies showed comparable MAE (IF, GroupDRO, ADV: 22.72, 22.46, 22.70 bpm) and similar fairness gaps (from 1.06 to 1.20 bpm). Compared to the unbalanced baseline (MAE: 24.74 bpm) from Table~\ref{tab:pretrain_vs_finetune}, these strategies offered modest improvements in accuracy (approx. 2bpm). In, BUT-PPG $\rightarrow$ MIMIC, GroupDRO attained the lowest MAE (20.14 bpm) but also the largest fairness gap (1.90 bpm). IF, in contrast, yielded a much lower gap (0.86 bpm) at the cost of a higher MAE (24.62 bpm). ADV's performance was intermediate (MAE, fairness gap: 24.92 bpm, 0.89 bpm).

\paragraph{Clinical-to-Consumer Adaptation: }
In MIMIC $\rightarrow$ DaLiA, IF substantially outperformed other methods (25.24 bpm vs. 42.15 bpm for GroupDRO and 36.54 bpm for ADV), though all methods achieved similar fairness gaps (ranging from 7.33 to 8.01 bpm). 
In MIMIC $\rightarrow$ BUT-PPG, ADV provided optimal accuracy (MAE: 27.21 bpm), while GroupDRO achieved the lowest demographic parity (fairness gap: 2.03 bpm) at the cost of higher prediction error.

\paragraph{Cross-Consumer Adaptation: }
In DaLiA $\rightarrow$ BUT-PPG, IF achieved the lowest MAE (13.44 bpm), while exhibiting a moderate fairness gap (1.30 bpm). Although GroupDRO (MAE, fairness gap: 15.47 bpm, 0.68 bpm) and ADV (14.49 bpm, 0.70 bpm) achieved better fairness than the unbalanced baseline (19.23 bpm, 2.36 bpm; see Table~\ref{tab:pretrain_vs_finetune}), they did so at the cost of higher predictive error.
In BUT-PPG $\rightarrow$ DaLiA, IF again demonstrated the best trade-off, with the lowest MAE (17.65 bpm) and largest gap reduction (6.86 bpm), while GroupDRO (MAE, fairness gap: 18.18 bpm, 7.12 bpm) and ADV (19.47 bpm, 7.15 bpm) performed comparably. These small gains suggest that the source and target domains are already well-aligned for transfers, limiting the benefits of explicit reweighting.

\vspace{-1ex}%Table 3. Latent Space Separability Metrics
\vspace{-3.0ex}
\begin{table}[ht]%[htbp]
\floatconts
  {tab:mmd_comparison}%
  {\caption{Effect of Bias Mitigation Strategies on Demographic Separability in Model Representations.}}%
  {%
    \small
    \begin{tabular}{l|c}
    \toprule
    \textbf{Method} & \textbf{MMD (Male vs. Female)} \\
    \midrule
    Unbalanced        & 0.004478 \\
    \midrule
    IF & 0.003896 \\
    GroupDRO         & 0.004354 \\
    ADV       & 0.004453 \\
    \bottomrule
    \end{tabular}
  }
\end{table}%\vspace{-1ex}
\subsection{Latent Representation Analysis}
Table~\ref{tab:mmd_comparison} reports the MMD between gender subgroups in the latent space across all datasets using stratified sampling. IF achieved the most significant reduction of MMD between gender subgroups by 13\% (from 0.004478 in the unbalanced setting to 0.003896), indicating reduced demographic separability in learned representations. GroupDRO and ADV showed modest reductions of 2.8\% and 0.5\%, respectively. 

%These reductions demonstrate that bias mitigation strategies can reorganize the model's internal feature representations, with IF showing the strongest effect in reducing gender-based clustering while maintaining physiological signal fidelity.

\vspace{-2.0ex}
\section{Discussion}
This study investigated the potential of bias-aware fine-tuning to improve both prediction accuracy and demographic fairness when adapting PPG foundation models for HR estimation across diverse clinical contexts. Fine-tuning emerged as a transformative step, yielding a 74–80\% error reduction across heterogeneous datasets. This underscores the critical role of cross-dataset adaptation in physiological AI, where signal characteristics vary markedly by device, patient population, and setting. Beyond improving accuracy, fine-tuning also restructured the model’s internal representations, transforming the general physiological knowledge encoded during pretraining into context-specific representations suited for local deployment—particularly valuable in settings with limited data and infrastructure.

%However, our findings challenge the common assumption that larger models invariably deliver better clinical performance. While increased model complexity sometimes improved overall accuracy, this improvement was not consistent across datasets or model sizes, particularly in cross-dataset settings, where larger models sometimes underperformed due to overfitting or fine-tuning instability. Moreover, model scaling consistently exacerbated demographic disparities, suggesting that larger models may be more susceptible to learning spurious correlations or demographic shortcuts embedded in training data. Detailed model scaling analysis across all dataset combinations is provided in Appendix \ref{apd:second}.
However, our findings challenge the assumption that larger models invariably deliver better clinical performance. While increased complexity occasionally improved accuracy, these gains were inconsistent across datasets and model sizes. In cross-dataset transfers, larger models sometimes underperformed likely due to overfitting or unstable fine-tuning. Moreover, scaling consistently exacerbated demographic disparities, suggesting susceptibility to spurious correlations or demographic shortcuts in the training data. 
%(Detailed model-scaling analysis across all dataset combinations is provided in Appendix \ref{apd:second}.) %In summary, bigger is not always better in this context – neither for accuracy nor for fairness.
%We evaluated PPG-GPT across multiple model sizes (19M to 1B parameters) to assess scaling effects on both accuracy and fairness. Based on this preliminary analysis, we selected the 19M parameter model for our comprehensive bias mitigation evaluation, as it provided a balanced foundation for systematic comparison across bias mitigation strategies. Complete model scaling analysis across all dataset combinations and parameter sizes is provided in Appendix \ref{apd:second}.
Divergent fairness patterns across transfer directions further underscore the context-specific nature of data and algorithmic bias: intra-category transfer (Cross-Consumer) showed different fairness dynamics compared to inter-category transfers (Consumer-to-Clinical and Clinical-to-Consumer), reflecting differences in demographic composition and signal acquisition conditions. These findings indicate that mitigation strategies must be tailored to deployment contexts rather than applied generically. Elevated cross-dataset errors largely reflected domain shift (see Results~\ref{results_sub1} and Appendix~\ref{apd:fourth} Table~\ref{tab:Intra_Bias_Mitigation_Performance}). %Notably, intra-dataset baseline results with the same datasets for train and test demonstrate that the elevated MAE scores observed in cross-dataset scenarios (17-35 bpm) primarily reflect domain shift challenges rather than inherent model limitations. 
Fairness gaps estimates also showed high variance across training runs in some scenarios (see Fig. \ref{fig:fairness_gap_unbalanced} bottom), indicating stability issues, especially when the test domain was small and demographically distinct (e.g., test set: BUT-PPG). Furthermore, reductions in MMD demonstrated that bias mitigation strategies can reorganize the model's internal feature space, with IF showing the strongest effect in reducing gender-based clustering while preserving physiological fidelity. 

Our evaluation of bias mitigation approaches demonstrates that fairness-aware fine-tuning can meaningfully reduce demographic disparities without sacrificing overall performance. IF weighting showed particular promise for wearable device scenarios, achieving both accuracy and fairness by counterbalancing demographic representation during training. GroupDRO demonstrated strength in clinical settings, achieving near-parity in some cases. By contrast, ADV yielded moderate gains but lacked consistency across transfer scenarios, likely due to its additional hyperparameter ($\lambda$) requiring dataset-specific tuning, as well as the inherent instability of balancing adversarial objectives with regression targets—demonstrating that algorithmic complexity does not guarantee superior fairness-performance trade-offs. 
Beyond model algorithms, real-world signal quality factors can also play a role: for instance, motion artifacts — often correlated with demographic and behavioral traits (\cite{chizari2024mitigation}) — may further exacerbate bias in PPG-based predictions. These signal-level disparities, rooted in device validation on demographically narrow cohorts, underscore the importance of considering data acquisition-level fairness. Furthermore, fairness gaps in physiological predictions may propagate into downstream clinical decisions, including alarm thresholds or early warning systems. Disparities in accuracy could result in miscalibrated alerts or delayed responses, disproportionately affecting specific demographic groups. This highlights the need for subgroup-aware calibration in real-world deployment of clinical AI models.

Despite the promising results, our study has several limitations. We focused primarily on binary gender disparities, chosen because it was recorded for all datasets, is conceptually straightforward, and served as an interpretable initial axis of demographic disparity — while other key attributes such as age, race/ethnicity, and socioeconomic status were not explored. Each of these factors involves distinct bias mechanisms (e.g., gender bias often stems from structural or social factors, whereas skin tone bias arises from sensor optics) and would require more complex modeling (multiclass or continuous). Moreover, such factors can interact with each other, complicating the isolation of any single bias sources. Additionally, dataset size and homogeneity may constrain generalizability to broader populations. Finally, some mitigation methods are computationally intensive—especially for larger models—which raises concerns about feasibility in resource-constrained settings. In future work, we aim to address these challenges by expanding demographic coverage, intersectional, and synthetic benchmarking, and by developing scalable debiasing strategies for deployment in diverse clinical environments.

\vspace{-3.0ex}
\section{Conclusion}
This study presents FairTune, a bias-aware fine-tuning framework for adapting the PPG-GPT foundation model to HR prediction tasks across heterogeneous datasets. While standard fine-tuning substantially improved accuracy (74--80\% error reduction), it failed to mitigate demographic disparities, potentially compromising equitable care. Our cross-dataset experiments revealed that scaling model size (up to 1B) often amplified gender-based gaps without consistent accuracy gains, challenging assumptions about fairness-performance trade-offs in large models. Importantly, FairTune effectively reduced disparities via lightweight mitigation: IF weighting showed most effective for consumer wearable data, and GroupDRO for clinical settings. These strategies fundamentally reorganize internal representations to emphasize physiological signals over demographic proxies, enabling more equitable HR prediction. These findings underscore that algorithmic bias manifests differently across domains, and that lightweight, bias-aware fine-tuning offers a practical solution for equitable model adaptation in healthcare settings where large-scale pretraining is infeasible.
%\clearpage} % preamble에 추가

\FloatBarrier
\bibliography{jmlr-sample}
\clearpage
\appendix

\renewcommand{\thetable}{A\arabic{table}}
\renewcommand{\thefigure}{A\arabic{figure}}
\setcounter{table}{0}
\setcounter{figure}{0}

\section{Mathematical Formulation of Bias Mitigation}\label{apd:first}
We implemented the following three bias-aware fine-tuning strategies:

\subsubsection{Inverse-Frequency Weighting (IF)}
Samples are reweighted inversely proportional to their demographic group frequency through weighted sampling. Let \( g(i) \) denote the demographic group of sample \( i \), and \( f_{g(i)} \) its empirical frequency. Each sample receives sampling weight:
\vspace{-1.5ex}
\begin{equation}
    w_i = \frac{1}{2 \cdot f_{g(i)}}
\end{equation}
where the factor of 2 normalizes weights across binary gender groups. Training uses WeightedRandomSampler with these weights while maintaining the standard MAE loss.

\subsubsection{Group Distributionally Robust Optimization (GroupDRO)}
Our GroupDRO implementation addresses group imbalance through adaptive weighted sampling based on group-specific performance. For groups \( \mathcal{G} \) with per-group losses \( \mathcal{L}_g \), we first normalize the group losses:
\vspace{-1.ex}
\begin{equation}
    \hat{\mathcal{L}}_g = \frac{\mathcal{L}_g - \min_{g' \in \mathcal{G}} \mathcal{L}_{g'}}{\max_{g' \in \mathcal{G}} \mathcal{L}_{g'} - \min_{g' \in \mathcal{G}} \mathcal{L}_{g'}}
\end{equation}
The group weights are then computed as:
\begin{equation}
    w_g = 1 + \eta \cdot \hat{\mathcal{L}}_g
\end{equation}
where \( \eta \) is a hyperparameter controlling the strength of reweighting. Each sample from group \( g \) receives sampling weight \( w_g / |\mathcal{G}_g| \), where \( |\mathcal{G}_g| \) is the group size. This approach upweights samples from under-performing groups during training through weighted random sampling.

\subsubsection{Adversarial Debiasing (ADV)}
ADV introduces an adversarial branch to discourage demographic leakage in latent representations. The training alternates between: (1) training the adversary to predict demographics, and (2) training the main model to maximize adversarial uncertainty:
\begin{equation}
    \mathcal{L}_{\text{total}} = \mathcal{L}_{\text{pred}}(y, \hat{y}) + \lambda \cdot \mathcal{H}(\hat{g})
\end{equation}
where \( \mathcal{H}(\hat{g}) = -\sum p_i \log p_i \) is the entropy of adversarial predictions, encouraging confusion in demographic classification.

\section{Maximum Mean Discrepancy Calculation Procedure}\label{apd:third}
MMD analysis employed stratified sampling to ensure balanced representation across datasets (DaLiA, BUT-PPG, MIMIC), gender groups (male/female), and heart rate bins centered around the median HR (83.1 bpm). For each stratum defined by dataset-gender-HR bin combinations, 50 samples were randomly selected when available, yielding 900 total samples for analysis.
Feature extraction utilized the model's penultimate layer representations after normalization and activation functions. For each bias mitigation strategy, features were extracted from models trained on corresponding datasets and concatenated across all three datasets to create unified male and female feature matrices.

MMD computation employed direct RBF kernel calculation:
$k(x,y) = \exp(-\gamma(\|x\|^2 + \|y\|^2 - 2\langle x,y \rangle))$, where $\gamma = 1.0$. 
Squared Euclidean distances were computed via matrix operations ($XX + YY - 2XY$) for computational efficiency. 
The final $\text{MMD}^2$ value was calculated as shown in Equation~\ref{eq:mmd_squared}:
\begin{equation}
\label{eq:mmd_squared}
\text{MMD}^2 = \frac{1}{n^2}\sum K_{xx} + \frac{1}{m^2}\sum K_{yy} - \frac{2}{nm}\sum K_{xy}
\end{equation}
where $n$ and $m$ represent male and female sample sizes respectively. 
This approach provides a robust measure of distributional differences in learned representations between gender groups while controlling for confounding factors through stratified sampling.

\section{Intra-Dataset Baseline Performance} \label{apd:fourth}
\begin{table*}[ht]
\floatconts
  {tab:Intra_Bias_Mitigation_Performance}%
  {\caption{Intra-Dataset Baseline Performance with Bias Mitigation Strategies for 19M PPG-GPT Model.}}%
  {%
    \small
    \begin{threeparttable}
    \resizebox{\linewidth}{!}{
    \begin{tabular}{ll|l|c|cc|c}
    \toprule
    \multirow{2}{*}{\textbf{Trainset}} & \multirow{2}{*}{\textbf{Testset}} & \multirow{2}{*}{\textbf{Method}} 
    & \multicolumn{3}{c|}{\textbf{Predictive Performance (MAE ↓)}} 
    & \textbf{Fairness Gap} \\
     & & & \textbf{Total} & \textbf{Male} & \textbf{Female} & \textbf{$|$Male$-$Female$|$} \\
    \midrule
    \multirow{4}{*}{DaLiA} & \multirow{4}{*}{DaLiA} & Unbalanced 
    & 4.28 [4.18--4.36] & 4.55 [4.47--4.70] & 4.06 [3.94--4.17] & 0.49 [0.41--0.63] \\
    \cmidrule{3-7}
    & & IF 
    & \textbf{4.28 [4.15--4.38]} & 4.45 [4.28--4.61] & 4.15 [4.02--4.25] & \textbf{0.30 [0.21--0.45]} \\
    & & GroupDRO 
    & 5.97 [5.79--6.19] & 6.20 [5.89--6.49] & 5.78 [5.58--5.94] & 0.42 [0.15--0.76] \\
    & & ADV 
    & 5.33 [5.12--5.73] & 5.80 [5.62--6.27] & 4.93 [4.70--5.28] & 0.87 [0.75--0.99] \\
    \midrule
    \multirow{4}{*}{BUT-PPG} & \multirow{4}{*}{BUT-PPG} & Unbalanced 
    & 9.94 [9.77--10.14] & 11.54 [11.11--11.73] & 8.40 [7.91--8.87] & 3.15 [2.24--3.81] \\
    \cmidrule{3-7}
    & & IF 
    & 9.94 [9.80--10.12] & 11.54 [11.09--11.74] & 8.40 [7.93--8.85] & 3.13 [2.26--3.81] \\
    & & GroupDRO 
    & 10.87 [10.65--10.96] & 10.96 [10.57--11.31] & 10.79 [10.21--11.33] & \textbf{0.18 [--0.77--0.90]} \\
    & & ADV 
    & \textbf{10.03 [9.90--10.22]} & 11.59 [11.06--12.01] & 8.54 [8.15--8.94] & 3.05 [2.26--3.86] \\
    \midrule
    \multirow{4}{*}{MIMIC} & \multirow{4}{*}{MIMIC} & Unbalanced 
    & 9.76 [9.46--10.00] & 11.40 [11.24--11.74] & 8.77 [8.08--9.22] & 2.62 [2.08--3.65] \\
    \cmidrule{3-7}
    & & IF 
    & 9.76 [9.38--10.28] & 11.43 [10.98--12.15] & 8.75 [8.18--9.61] & 2.68 [1.77--3.88] \\
    & & GroupDRO 
    & 11.37 [10.24--12.31] & 12.19 [11.59--12.41] & 10.87 [9.42--12.31] & \textbf{1.32 [0.01--2.48]} \\
    & & ADV 
    & \textbf{10.16 [10.01--10.27]} & 11.69 [11.10--12.58] & 9.23 [8.70--9.77] & 2.46 [1.33--3.88] \\
    \bottomrule
    \end{tabular}
    }
    \begin{minipage}{\linewidth}
    \scriptsize
    Values are reported as median [95\% confidence interval] unless otherwise indicated. IF = Instance-weighted Fairness (Balanced training approach). Best performance values are shown in \textbf{bold} for each dataset pair.
    \end{minipage}
    \end{threeparttable}
  }
\end{table*}
Table~\ref{tab:Intra_Bias_Mitigation_Performance} presents intra-dataset baseline results for completeness. When models were trained and tested on the same dataset, MAE scores were much lower (DaLiA: 4.28 bpm, BUT-PPG: 9.94 bpm, MIMIC: 9.76 bpm). These baselines serve as achievable lower bounds under matched train–test conditions and indicate that elevated cross-dataset evaluation errors primarily reflect domain shift rather than inherent model limitations.

Within these baseline settings, bias mitigation strategies also proved effective in reducing subgroup disparities. In DaLiA $\rightarrow$ DaLiA, IF achieved the best fairness improvement, reducing the gap from 0.49 to 0.30 bpm while maintaining accuracy (MAE: 4.28 bpm). For BUT-PPG $\rightarrow$ BUT-PPG, GroupDRO dramatically reduced the fairness gap from 3.15 to 0.18 bpm with minimal accuracy cost (MAE: 10.87 vs. 9.94 bpm). In MIMIC $\rightarrow$ MIMIC, GroupDRO again showed superior fairness performance, reducing the gap from 2.62 to 1.32 bpm.

\end{document}